\documentclass[twoside,11pt]{article}

\usepackage{jmlr2e}
\usepackage{lastpage}

\jmlrheading{23}{2022}{1-\pageref{LastPage}}{9/21; Revised
5/22}{8/22}{21-1127}{Jiayi Weng, Huayu Chen, Dong Yan, Kaichao You, Alexis Duburcq, Minghao Zhang, Yi Su, Hang Su and Jun Zhu}
\ShortHeadings{Tianshou: A Highly Modularized Deep Reinforcement Learning Library}{Weng, Chen, Yan, You, Duburcq, Zhang, Su, Su and Zhu}
\firstpageno{1}

\begin{document}

\title{Tianshou: A Highly Modularized Deep Reinforcement Learning Library}

\author{ \\
\name Jiayi Weng$^\ddag$\thanks{J. Weng and H. Chen contributed equally to this article.} \email trinkle23897@gmail.com \\
\name Huayu Chen$^\ddag$\footnotemark[1] \email chenhuay21@mails.tsinghua.edu.cn \\
\name Dong Yan$^\ddag$ \email sproblvem@gmail.com \\
\name Kaichao You$^\S$ \email youkaichao@gmail.com\\
\name Alexis Duburcq$^\P$ \email alexis.duburcq@gmail.com\\
\name Minghao Zhang$^\S$ \email mehoozhang@gmail.com\\
\name Yi Su$^\sharp$ \email nuance@gmail.com\\
\name Hang Su$^\ddag$\thanks{H. Su and J. Zhu are corresponding authors.\vspace{-0.55cm}} \email suhangss@tsinghua.edu.cn \\
\name Jun Zhu$^\ddag$\footnotemark[2] \email dcszj@tsinghua.edu.cn \\
\addr $^\ddag$Dept. of Comp. Sci. \& Tech., BNRist Center, Institute for AI, Tsinghua-Bosch Joint ML Center, THBI Lab, Tsinghua University, Beijing, 100084, China \\
\addr $^\S$School of Software, Tsinghua University, Beijing, 100084, China \\
\addr $^\P$Wandercraft, 88 Rue de Rivoli, Paris, 75004, France\\
\addr $^\sharp$Ant Group, 525 Almanor Ave, Sunnyvale, CA, 94085, United States of America}
\editor{Antti Honkela}

\maketitle

\begin{abstract}%   <- trailing '%' for backward compatibility of .sty file
In this paper, we present \texttt{Tianshou}, a highly modularized Python library for deep reinforcement learning (DRL) that uses PyTorch as its backend.
Tianshou intends to be research-friendly by providing a flexible and reliable infrastructure of DRL algorithms.
It supports online and offline training with more than 20 classic algorithms through a unified interface.
To facilitate related research and prove Tianshou's reliability, we have released Tianshou's benchmark of MuJoCo environments, covering eight classic algorithms with state-of-the-art performance.
We open-sourced Tianshou at \url{https://github.com/thu-ml/tianshou/}.
\end{abstract}

\begin{keywords}
  Deep Reinforcement Learning, Library, PyTorch, Modularized, Benchmark
\end{keywords}

\section{Introduction}

Recent advances in deep reinforcement learning (DRL) have ignited enthusiasm of both academia and industry. This is accompanied by the flourish of many newly-emerged DRL algorithms \citep{mnih2015human, alphagozero, duan2016benchmarking}, together with numerous libraries that try to provide reference implementations with the representative ones including \texttt{RLlib} \citep{rllib}, \texttt{rlpyt} \citep{stooke2019rlpyt}, \texttt{Stable-Baselines3} \citep{stable-baselines3}, \texttt{MushroomRL} \citep{d2020mushroomrl}, and \texttt{PFRL} \citep{fujita2021chainerrl}.

Though most DRL libraries are comprehensive, many researchers still tend to use their own DRL code base for fast prototyping in practice. The reasons behind this are various. Some libraries may choose to highly encapsulate the supported algorithms, leaving out several options to be tweaked. This assists algorithms' application, but harms the flexibility because it is impossible to provide exhaustive options. The usability is also a concern. Several libraries prioritize supporting highly distributed training, but bring complex code structure and difficulties in debugging. However, parallelized data sampling is still needed in research because of the typically unbalanced numbers of CPUs and GPUs in a server. Another reason might be that many libraries are only comprehensive for a certain type of algorithms (e.g., only support online or offline algorithms) to keep a unified API.

To address these issues, we present Tianshou, a highly modularized Python library for deep reinforcement learning based on PyTorch. Tianshou has the following characteristics:
\begin{itemize}
\item  \textbf{Highly modularized.}
Tianshou aims to provide building blocks rather than training scripts.
It can be easily used for fast prototyping because the shared infrastructure commonly used in DRL is factored out (Figure \ref{fig:overview}).
Users only need to change a few variables to apply techniques commonly used in DRL (e.g., parallel data sampling).

\item \textbf{Reliable.}
Tianshou has a code coverage of 94\%. Every commit to Tianshou will go through unit tests on multiple platforms. We have also released a systematic benchmark of Gym's MuJoCo environment\footnote{\url{https://tianshou.readthedocs.io/en/master/tutorials/benchmark.html}} \citep{todorov2012mujoco, brockman2016openai}. In this benchmark, Tianshou incorporates a comprehensive set of DRL techniques for 8 benchmarked algorithms and scores 15\% higher on average in terms of median performance compared with reference implementations.

\item \textbf{Comprehensive.} Besides comprehensive model-free algorithms, Tianshou also supports offline learning and many other DRL techniques such as GAIL \citep{gail} and ICM \citep{icm}. Moreover, through a unified Python interface, Tianshou formulates the data collecting (e.g., both synchronous and asynchronous environment execution) and agent training paradigms in DRL. Lastly,
Tianshou has plentiful functionalities that may extend its application (see Section \ref{Architecture}).

\end{itemize}

\section{Architecture of Tianshou}
\label{Architecture}
\begin{figure}[t]
    \centering
    \includegraphics[width=1.0\linewidth]{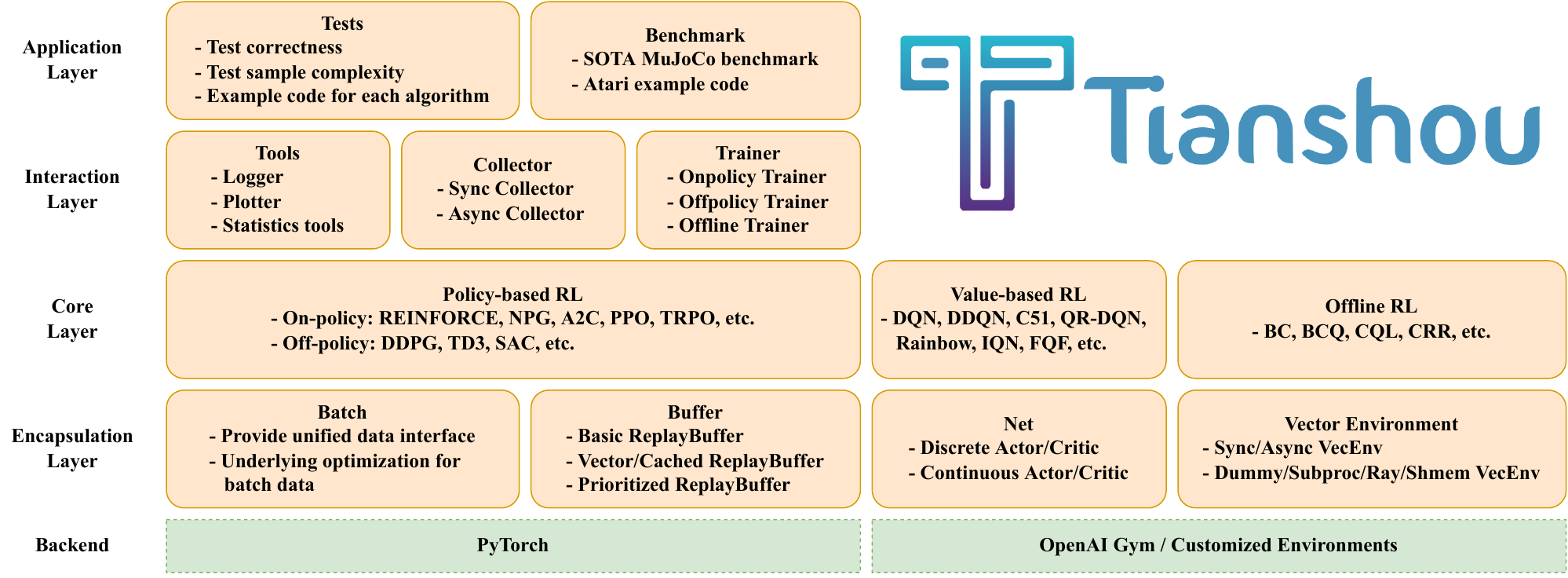}
    \caption{Tianshou's architecture. Tianshou consists of 4 layers: the bottom-level encapsulation layer provides encapsulations of third-party libraries; the core layer implements many kinds of DRL algorithms; the interaction layer aims to offer user-oriented high-level APIs; and finally, the application layer provides training scripts to demonstrate Tianshou's usage. Instances of the same building block share almost the same pythonic APIs.
    }
    \label{fig:overview}
\end{figure}

In this section, we will briefly introduce Tianshou's architecture as illustrated in Figure~\ref{fig:overview}.
\begin{figure}[t]
    \centering
    \includegraphics[width=0.7\linewidth]{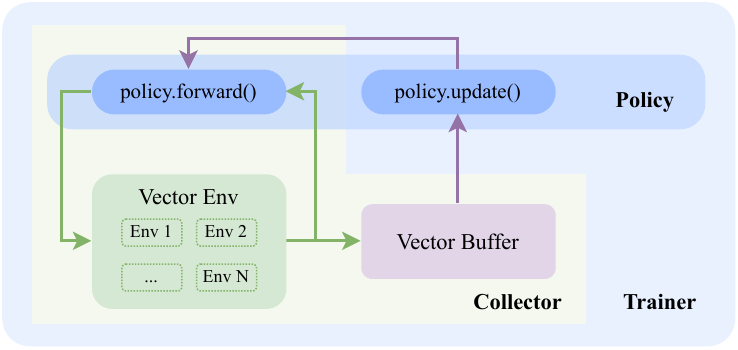}
    \caption{A high-level depiction of Tianshou's training process.}
    \label{fig:parallel}
\end{figure}

\textbf{Standardization of the Training Process.} 
% We have standardized and categorized the training paradigms of mainstream DRL algorithms into three types by considering different mechanisms of experience replay: 
We standardize the training paradigms of mainstream DRL algorithms by considering different experience replay mechanisms and classify them into three types:
on-policy training, off-policy training, and offline learning.
We use a replay buffer to store the transitions and a collector to collect transition data into the buffer. We use the policy's \texttt{update} function to update the parameter (Figure~\ref{fig:parallel}).

\textbf{Parallel Computing Infrastructure.} In concurrent research, \citet{stooke2019rlpyt} addresses two phases of parallelization in DRL: environment sampling and agent training. Tianshou targets small- to medium-scale research, so it focuses on the first one. Following \citet{clemente2017efficient}, we adopt their parallelization technique to balance simulation and inference loads. Note that our contributions to parallel sampling schemes exceed this work by allowing asynchronized sampling as an alternative way to ease the straggler effect instead of only stacking environment instances per process. Thanks to the modularized design, Tianshou can easily support the C++-based vectorized environment EnvPool \citep{envpool} with free speed up.

\textbf{Utilities.} Tianshou intends to relieve users from imperceptible details critical to a desirable performance and, hence, incorporates a comprehensive set of DRL techniques as its infrastructure. Techniques include partial-episode bootstrapping \citep{pmlr-v80-pardo18a}, observation/value normalization \citep{van2016learning}, automatic action scaling and GAE \citep{gae}, etc.
Besides, Tianshou has plentiful extra functionalities that users might find helpful. For instance, Tianshou has customizable loggers compatible with TensorBoard and W\&B. Recurrent state representation, prioritized experience replay, training resumption, and buffer serialization are also supported. 

\textbf{Reproduction Scripts and Performance.} We have released Tianshou's OpenAI Gym MuJoCo task suite benchmark, covering 8 classic algorithms and 9 environments.
In this benchmark, Tianshou scores 15\% higher on average compared with multiple reference implementations in terms of 9 environments' median performance, demonstrating its reliability. For discrete action space problems, we also provide example code and results with 7 supported algorithms in 7 Atari environments. All experiments are done with 10 random seeds. Some libraries \citep{fujita2021chainerrl} devote themselves to faithfully replicating existing papers, while Tianshou aims to present an as-consistent-as-possible set of hyperparameters and low-level designs. While leaving the core algorithm untouched, we try to incorporate several known tricks in a specific algorithm to all similar algorithms supported by Tianshou. Hopefully, this will facilitate comparisons between algorithms.

\section{Usability}
Tianshou is lightweight and easy to install. Users can simply install Tianshou via Pip or Conda on different platforms (Windows, macOS, Linux). Full API documentation and a series of tutorials are provided at \url{https://tianshou.readthedocs.io/}. Only a few lines of code are required to start a simple experiment. 
Tianshou also strictly follows the PEP8 code style with the code commented and data type annotated. Contributing guidelines and extensive unit tests with GitHub Actions, including code-style, type, and performance checks, help Tianshou maintain its code quality.

\section{Comparison to Related Works}
While several TensorFlow-based DRL libraries \citep{tensorforce, plappert2016kerasrl, rlcoach} are also worth mentioning, we limit our comparison to a few DRL libraries with the PyTorch backend due to page limit. \texttt{RLlib} \citep{rllib} and \texttt{rlpyt} \citep{stooke2019rlpyt} are libraries designed to be high-throughput software and support both multi-CPU parallel sampling and multi-GPU optimization, while \texttt{Stable-Baselines3} \citep{stable-baselines3}, \texttt{PFRL} \citep{fujita2021chainerrl} and Tianshou focus on small- to medium-scale application of DRL algorithms and support only parallel sampling. Libraries like \texttt{MushroomRL} \citep{d2020mushroomrl} are intentionally designed to be research-friendly, so no parallelization is supported. This leads to very different design choices and results in different code complexity. In terms of supported algorithms, most libraries are comprehensive, but each has a different focus. \texttt{MushroomRL} supports both classic RL and DRL algorithms to facilitate research. \texttt{d3rlpy} \citep{seno2021d3rlpy} prioritizes supporting offline DRL algorithms and other libraries mainly support online algorithms. While many libraries above are modular, Tianshou achieves modularity mainly by factoring out the infrastructure in DRL, compared to several libraries that focus on offering one highly encapsulated API for each algorithm \citep{stable-baselines3, seno2021d3rlpy}. Above all, the architecture of \texttt{PFRL} is most similar to Tianshou. However, they still have key differences like the implementation of lower-level data container Batch and how to support sequence Buffers for RNN. Other engineering features are compared in Tianshou's GitHub repository\footnote{\url{https://github.com/thu-ml/tianshou/blob/master/README.md\#why-tianshou}}.

\section{Conclusion}
This paper briefly describes Tianshou, a flexible and reliable implementation of a modular DRL library. Tianshou sets up a framework for DRL research by factoring out the shared infrastructure commonly used in DRL as building blocks. We have also released a MuJoCo benchmark, covering many classic algorithms, demonstrating Tianshou's reliability.

% Acknowledgements should go at the end, before appendices and references

\acks{We thank Haosheng Zou for his early work on TensorFlow-based Tianshou before version 0.1.1. We thank Peng Zhong, Qiang He, Chengqi Duan, Qing Xiao, Qifan Li, Yan Li, and others for their valuable contributions to Tianshou.

This work was supported by the National Key Research and Development Program of China (
2020AAA0106000, 2020AAA0104304, 2020AAA0106302, 2021YFB2701000), NSFC Projects (Nos. 62061136001, 62076147, U19B2034, U1811461, U19A2081, 61972224), Beijing NSF Project (No. JQ19016), BNRist (BNR2022RC01006), Tsinghua Institute for Guo Qiang, Beijing Academy of Artificial Intelligence (BAAI), Tsinghua-Huawei Joint Research Program, and the High Performance Computing Center, Tsinghua University.
}

\vskip 0.2in
\bibliography{sample}

\end{document}